\documentclass[10pt,twocolumn,letterpaper]{article}

\usepackage{cvpr}              

\usepackage{graphicx}
\usepackage{amsmath}
\usepackage{amssymb}
\usepackage{booktabs}
\usepackage{multirow}
\usepackage{booktabs}
\usepackage{colortbl}
\usepackage[accsupp]{axessibility}

\usepackage[pagebackref,breaklinks,colorlinks]{hyperref}

\usepackage[capitalize]{cleveref}
\crefname{section}{Sec.}{Secs.}
\Crefname{section}{Section}{Sections}
\Crefname{table}{Table}{Tables}
\crefname{table}{Tab.}{Tabs.}

\def\cvprPaperID{*****} 
\def\confName{CVPR}
\def\confYear{2022}

\begin{document}

\title{Multi-encoder Network for Parameter Reduction of a Kernel-based Interpolation Architecture}

\author{Issa Khalifeh$^{1,2}$, Marc Gorriz Blanch$^{1}$, Ebroul Izquierdo${^2}$, Marta Mrak${^1}$\\
$^{1}$British Broadcasting Corporation, London, W12 7TQ \\
$^{2}$Queen Mary University of London, London E1 4NS\\
\tt\small \{i.khalifeh,ebroul.izquierdo\}@qmul.ac.uk \\
\tt\small \{marc.gorrizblanch, marta.mrak\}@bbc.co.uk} 
\maketitle
\begin{abstract}
Video frame interpolation involves the synthesis of new frames from existing ones. Convolutional neural networks (CNNs) have been at the forefront of the recent advances in this field. One popular CNN-based approach involves the application of generated kernels to the input frames to obtain an interpolated frame. Despite all the benefits interpolation methods offer, many of these networks require a lot of parameters, with more parameters meaning a heavier computational burden. Reducing the size of the model typically impacts performance negatively. This paper presents a method for parameter reduction for a popular flow-less kernel-based network (Adaptive Collaboration of Flows). Through our technique of removing the layers that require the most parameters and replacing them with smaller encoders, we reduce the number of parameters of the network and even achieve better performance compared to the original method. This is achieved by deploying rotation to force each individual encoder to learn different features from the input images. Ablations are conducted to justify design choices and an evaluation on how our method performs on full-length videos is presented.

\end{abstract}

\section{Introduction}
\label{sec:intro}

Video frame interpolation is the process of generating an intermediate frame from a set of input frames. It is used in a wide range of applications such as slow-motion video generation, adapting old content for modern high frame-rate TVs, the replacement of defective frames and the reduction of jitter.

Traditionally, the optical flow is used to obtain the motion vectors between the input frames. This motion information is used to warp the input image to obtain the interpolated image before undergoing post-processing. Artefacts are sometimes present in the final output and are usually caused by an incorrect motion vector calculation. Blurriness, sudden brightness changes and occlusions in the input frames are among the main contributors to inaccurate calculations.

With deep learning, the resilience of optical flow-based methods has improved. As a result, more complex motions can be better handled. This is evidenced by the favourable performance of recent methods such as BMBC \cite{park2020bmbc}. To avoid relying on optical flow, the SepConv \cite{niklaus2017video} method was devised. SepConv combines the process of finding the motion between frames and the process of warping into a single step with these processes being done implicitly. In such an approach, the output of the UNet is used as an input to four subnet blocks, producing horizontal and vertical kernels for each input image. These kernels are convolved with each input image. The convolved images are then summed together to obtain the interpolated frame.

The success of this method has led to the emergence of other variants of this work. One such variant is Adaptive Collaboration of Flows (AdaCoF) \cite{lee2020adacof} which addresses memory usage issues as well as improving overall performance. Better memory usage is achieved through the use of smaller more flexible kernels. In contrast to SepConv which uses a fixed kernel, AdaCoF uses deformable offsets which allow for the better referencing of pixels when interpolating a frame. Occlusion maps are used to better handle occlusion cases and represents an explicit handling of this special case.

Although such a method allows for greater flexibility in referencing pixels, it still uses a considerable amount of parameters, approximately 21.4 million. Performance is usually negatively impacted if steps were taken to reduce parameters. 

In the design process of our network, it was found that the final two convolution blocks in the 5-level (512 channel) encoder and the first two decoder blocks account for 91$\%$ of total parameters. Removing these blocks negatively impacted performance on some of the evaluation sets. This was due to the loss of the deeper, more intrinsic features that are typically obtained by deeper networks. The original method had feature maps of up to 512 channels whereas, after the removal of these blocks, the resulting network has 128 channels. To compensate for this, a multi-encoder approach was devised. Each encoder focuses on separate features within the input images and we demonstrate the differences in learning by visualising the occlusion and attention maps for our model.
\begin{figure*}
\centering
\begin{center}
   \includegraphics[width=0.9\linewidth]{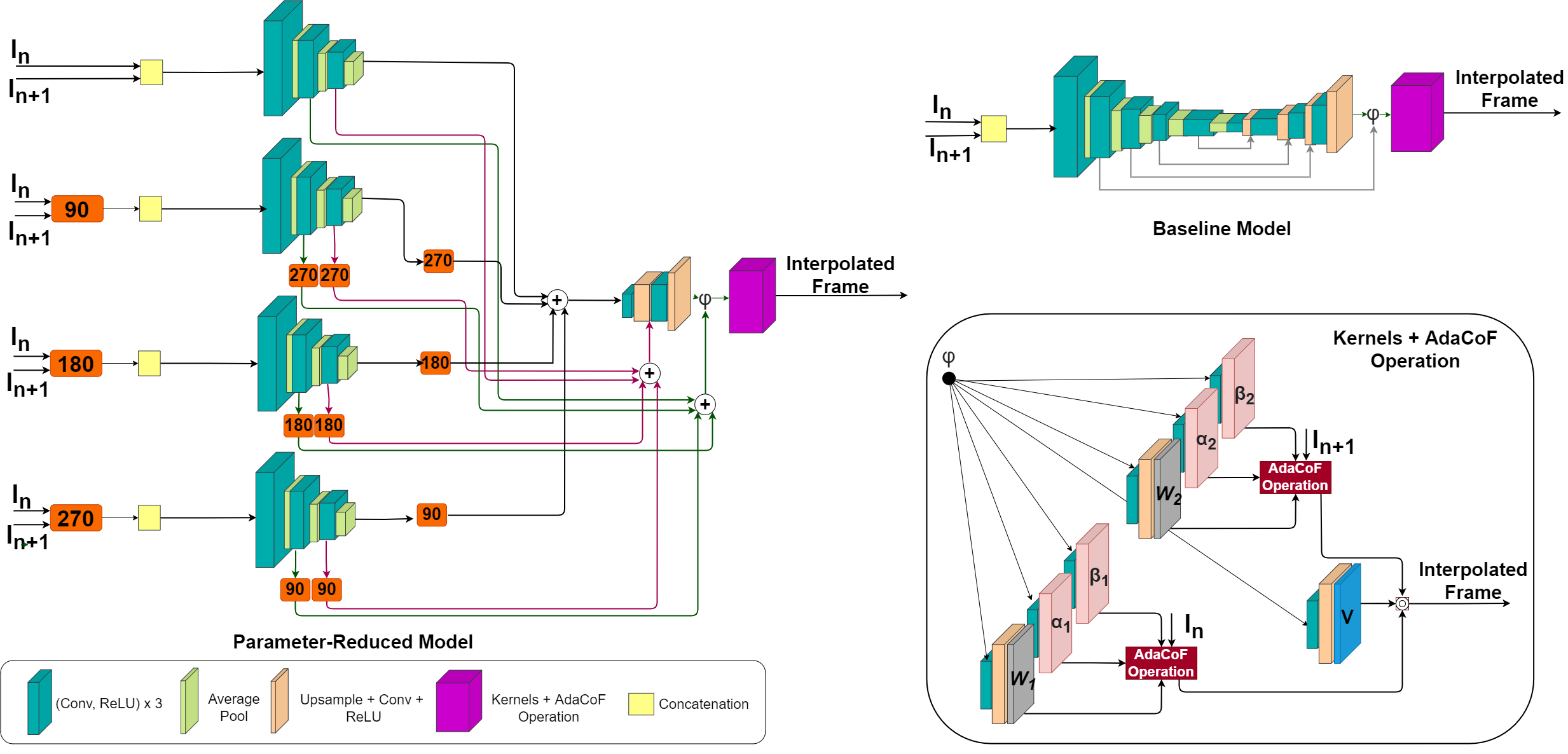}
\end{center}
   \caption{A representation of the proposed network architecture. Each encoder input is rotated by $0^{\circ}$, $90^{\circ}$, $180^{\circ}$, $270^{\circ}$ depending on the encoder number. The output feature maps at each level are rotated to ensure they are all the same orientation ($0^{\circ}$) before being added together. This combined output is then used as a skip connection input to the decoder at each corresponding level. }
\label{fig:long}
\label{fig:legacy}
\end{figure*}

The multi-encoder approach deployed here is modular and simply involves using 4 encoders with the same architecture. The training procedure is straightforward and not so different from the training procedure of the original AdaCoFNet method which we use as a base to simplify. This means that, these changes can be easily implemented without requiring different frameworks to achieve optimised performance. 

In this paper, we propose a simplified model that make use of our multi-encoder approach. Our network uses 84.5$\%$ less parameters compared to the baseline whilst maintaining a similar level of performance. This network is termed \textbf{P}arameter-\textbf{R}educed \textbf{Net}work for Video Frame Interpolation and will be referred to as $PRNet_{L}$ in the paper, with $L$ being the total number of encoders in the network.

Therefore, the contributions of this paper are the following:

\textbf{$-$} We propose a modular multi-encoder approach which compensates for the loss of deeper, more intrinsic features by leveraging the power of multiple encoders

\textbf{-} The total number of parameters of the network can be reduced by 84.5$\%$ whilst being able to achieve similar performance on the evaluation sets

\textbf{-} Our method makes use of rotations which allows the network to learn different representations of the input image and enables the network to achieve a better overall performance
\section{Related Work}
\label{sec:related}
The growth of CNNs has enabled the creation of new methods to handle the problem of video frame interpolation. One of the first to approach this problem  was \cite{long2016learning} which introduced a direct synthesis network with no optical flow dependency. The results were blurry and were not a major improvement for optical flow methods. It did, however, serve as a proof of concept which could enable the development of CNNs for frame interpolation. Methods Super-Slomo \cite{jiang2018super} integrated an optical flow CNN and conducted flow refinement to better handle interpolation artefacts.

The concept of using kernels was first introduced by \cite{niklauscvpr2017}. As it removed the need for explicit motion-based interpolation, this method provided a good alternative to direct synthesis networks such as \cite{long2016learning} or using optical flow networks. The advantages of this method meant that, as with \cite{long2016learning}, videos taken in the wild could be used for training allowing for a wider range of scenarios to be accounted for in the dataset. This would be especially helpful in reducing the domain gap as optical flow CNN models are trained using synthetic datasets such as KITTI \cite{Geiger2012CVPR} where the flow maps are present and can be used as ground truth. 

Newer methods have refined the initial model introduced by Niklaus et al. One such example is SepConv \cite{niklauscvpr2017} which significantly reduces memory usage by using separable kernels. Despite the reduction in memory usage, memory was still being used inefficiently due to the use of fixed kernels (set to a size of 51). For example, if there was a pixel movement of 5 pixels, a kernel size of 51 would not be required and a smaller kernel size would perform well. In addition to AdaCoF, there are other approaches which tackle this problem such as DSepConv\cite{cheng2020video} which introduces deformable separable convolutions and EDSC \cite{cheng2020multiple} which deploys a similar approach but presents other works as examples of their generalised method. 

Other models have tried combining the advantages of both methods. Examples include DAIN \cite{bao2019depth} and MEMC-NET \cite{bao2019memc} which combine flow and kernel-based methods. DAIN integrates context, depth, flow-estimation and kernel-estimation networks. Despite this, the limitations in terms of pixel referencing still remain. This, however, does not mean that optical flow-based methods have lost their popularity. Methods such as SoftSplat \cite{niklaus2020softmax} and BMBC \cite{park2020bmbc} perform very well in interpolation benchmarks and demonstrate that interpolation methods can be better designed to handle issues with optical flow.
One field that hasn't received much attention due to the difficultly in achieving visually pleasing results is direct synthesis networks. Such networks don't make use of kernels (to extract weights for a special convolution operation) or optical flow (to extract motion vectors) and simply the output of the network is the interpolated frame. As mentioned earlier \cite{long2016learning} did indeed produce a direct synthesis network and although the development was significant at the time, compared to traditional methods there was still a long way to go. \cite{choi2020channel} introduced the concept of using channel-attention with residual blocks \cite{zhang2018image} for interpolation. In addition to introducing this concept to interpolation, pixelshuffle \cite{shi2016real} is made use of for the downsampling operation. The results of this are promising for the future of direct synthesis interpolation networks.

Recently, 3D CNNs have been making headway in the frame interpolation field. \cite{kalluri2020flavr} first introduced the concept of using a 3D UNet for interpolation. Though they were indeed pioneers for introducing the community to the benefits of frame interpolation and the potential of 3D CNNs, there are issues when it comes to training time. The authors train on 8 GPUs with a batch size of 32. On a single 2080Ti GPU with a batch size 4 (maximum possible to use all available VRAM), training takes around 2.5 months. Additionally the network uses the larger Vimeo90K Septuplet set which is larger than the triplet set and thus the read time is longer compared to the smaller Vimeo Triplet dataset \cite{xue2019video}. Although the concept is indeed interesting and is a direct synthesis network, the network uses 4 input images which might be a constraint when a user simply doesn't have enough frames to run the algorithm. 
There have also been new developments integrating 3D CNNs with transformers. \cite{shi2021video} adapted the Swin transformer \cite{liu2021swin} to take into consideration the temporal dimension. The results look very promising and show potential in the interpolation field.

This paper focuses on reducing the parameters of a popular 2D approach named Adaptive Collaboration of Flows \cite{lee2020adacof}. The reason for the selection of this paper lies with the ease of training, the fairly short training duration (approximately 5 days) and the flexibility of this method in running on different environments meaning it would be better accessed by wider audiences.

The method for parameter reduction is inspired by work in the deep image compositing field \cite{zhang2021deep}, whereby a multi-encoder architecture is used to fuse the foreground and background images effectively. The aim is to obtain an output image where these two images are fused seamlessly. The rationale behind transferring this multi-encoder approach for interpolation mainly stems from the deduction that each encoder extracts various features from the input images and that could complement each other and could in theory be used to compensate for the lack of deeper, more intrinsic features that are usually obtained with deeper CNNs.This paper highlights the benefits of such an approach and also supports conclusions through ablations and test on full-length 100\% shutter sequences with a range of different non-linear motions which tend to be challenging in an interpolation context.
\begin{figure*}
\centering
\begin{center}
   \includegraphics[width=0.9\linewidth]{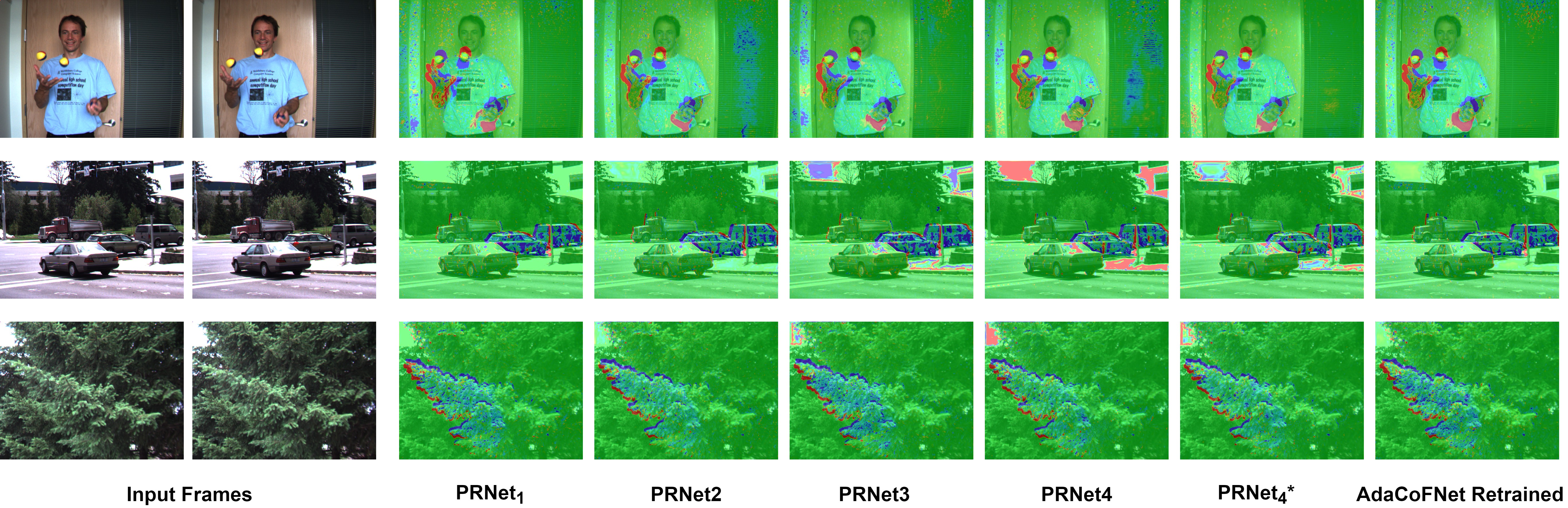}
\end{center}
   \caption{A visualisation of the Occlusion maps for the models used in the ablation. }
\label{fig2:long}
\label{fig2:legacy}
\end{figure*}
\section{Proposed Approach}
\subsection{Base Method}

This paper selects AdaCoFNet method as the baseline interpolation method. It is a kernel-based method which improves the handling of complex motions by introducing occlusion reasoning as well as deformable offsets, thereby increasing the degrees of freedom of previous video frame interpolation approaches.

Given two consecutive video frames $I_{n}$ and $I_{n+1}$, the intermediate frame $I_{out}$ is predicted by warping both input images via operation $\mathcal{T}$ as follows:
\begin{equation}
I_{out} = V \odot \mathcal{T}_f(I_{n}) + (K-V) \odot \mathcal{T}_{b}(I_{n+1}),
\label{eq1}
\end{equation}
where $V$ is the occlusion reasoning, $K$ is a matrix of ones and $\odot$ is pixel-wise multiplication. $\mathcal{T}_{f}$ and $\mathcal{T}_{b}$ denote forward and backwards warping. 

In particular, the AdaCoF operation can be used to approximate the $\mathcal{T}$ warping. In this operation, the adaptive kernel weights $W_{k, l}$ and offset vectors from each output pixel $(\alpha_{k,l} , \beta_{k,l})$ are applied to each input image. In applying this operation, an approximation of the forward and backward warped output is obtained. These due outputs are then weighted according to the Occlusion map $V$. Occlusion reasoning is used to handle the case where a pixel is only present in one of the input images. The image where the pixel is visible is weighted more heavily than the one where it is not. These kernels $W_{k, l}$, $(\alpha_{k,l}$ and $\beta_{k,l})$ as well as how they relate to the AdaCoF operation are presented in Figure 1.

In order for the inputs to these operations to be obtained, a UNet architecture \cite{ronneberger2015u} is used and this is shown in the top right corner of Figure 1 as part of the baseline model. The two input images are stacked in the channel dimension and used as an input to the network. The combined feature map output from the second encoder block and the final decoder block, named $\psi$ in Figure 1 goes into 7 subnets required for this operation. The subnets then estimate the AdaCoF parameters $(W_{k, l}, \alpha_{k,l} , \beta_{k,l})$ for each frame and the occlusion mapping $V$.

Being $\hat{I}$ the warped frame from $I$, AdaCoF operation can be summarised as follows:
\begin{equation}
\hat{I}(i, j) = \sum_{k=0}^{F-1}\sum_{l=0}^{F-1} W_{k, l}(i,j) I(i+dk+\alpha_{k,l}, j+dl+\beta_{k,l}),
\label{eq3}
\end{equation}
\noindent where $F$ is the kernel size and $d \in \{0, 1, ...\}$ is a dilation factor used as starting point for the offset values. All the experiments in this paper use dilation $d=1$ and kernel size $F=5$.
The subnets and the AdaCoF operation are shown in Figure 1.
\subsection{Proposed Multi-Encoder Approach}
Multi-encoder approaches have been used in different fields for a wide range of purposes, notably in deep image compositing \cite{zhang2021deep} where the inspiration for our work comes from. In deep image compositing, two encoders are used to fuse the features of the background and foreground images together. The difference in our approach lies in the fact that we deploy a multi-encoder architecture for the purpose of parameter reduction and not for fusing different images. A similar rationale applies.

The UNet backbone of AdaCoF consists of 5 encoder convolutional blocks or levels L, with each convolutional block containing 3 convolutions and 3 ReLU activations, please refer to Figure 1 for more details. The output at encoder levels L={1,2,3,4,5} consists of feature maps with {32,64,128,256,512} channels. The fourth and fifth encoder blocks (256 and 512 channel outputs) and the first two decoder and upsample blocks (512 and 256 channel outputs) utilise the highest proportion of parameters of the network, approximately 91$\%$ of all total trainable parameters. By removing these blocks, the resultant UNet is 3-level, meaning output feature maps at L={1,2,3} are {32,64,128}. However, the removal of these deep intrinsic features results in a drop in performance as the network is now less able to handle the complex motions present in real-world sequences.

To compensate for the removal of these intrinsic features, additional 3 level encoder blocks are added to the network architecture and the output at each level for each encoder is combined before being passed as a skip connection to the decoder. The rationale behind this is that each encoder can specialise by focusing on different features for each input image. Each encoder could complement the other, breaking down the problem into more manageable steps. This is necessary in a shallow 3-level encoder architecture which lacks the deeper features of the 5-level encoder UNet.

\begin{figure*}
\centering
\begin{center}
   \includegraphics[width=0.8\linewidth]{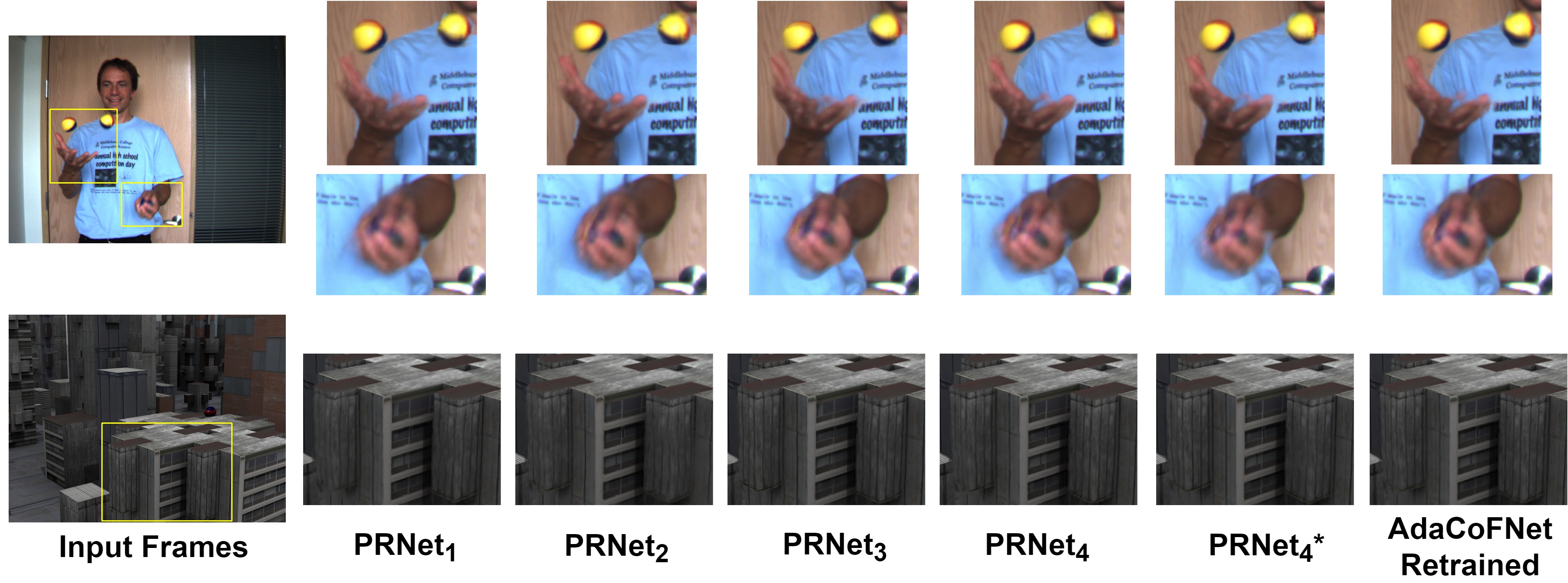}
\end{center}
   \caption{The output of the different PRNet networks with a particular highlight on tricky regions }
\label{fig4:long}
\label{fig4:legacy}
\end{figure*}
Rotations are also an important tool to better handle complex motions as they allow the network to learn different representations of the inputs. The performance of a model on certain sequences may be better when rotation is involved. For example, using our retrained version of AdaCoFNet, performance on the Middlebury Other set is 35.79dB. When the inputs (and then output) are rotated by $90^{\circ}$, performance increases to 35.84dB albeit performance on the UCF-101 and DAVIS sets drops. Rotations by $180^{\circ}$ and $270^{\circ}$ result in a degradation in performance on all the sets used in this paper. This indicates that the applications of rotation are limited and specific to certain input sequences. One reason for this change in performance is that the network might be biased towards some directions of motion (e.g. horizontal) and that by rotating the input sequence, the network performs better.

To better equip the network for a wide range of motions, rotations are applied to the input frames $I_{n}$ and $I_{n+1}$, of the proposed 4 encoder network. The input to each encoder would be rotated by a different angle. Encoders 1, 2, 3 and 4 would have the inputs rotated by $0^{\circ}$ (no rotation), $90^{\circ}$, $180^{\circ}$, $270^{\circ}$. 
As mentioned earlier, the output feature maps $\phi_{L}$ produced at level L where L={1,2,3} are combined before being passed as a skip connection. In the case of each input sequence being rotated, all feature maps obtained from each encoder are rotated again so that they are the same orientation. For inputs rotated by $0^{\circ}$ (no rotation), $90^{\circ}$, $180^{\circ}$, $270^{\circ}$ , the feature maps would be rotated by $360^{\circ}$ (no rotation), $270^{\circ}$, $180^{\circ}$, $90^{\circ}$. This is represented in Equation 3

\begin{equation}
\theta_{L}^{N} = \sum_{k=0}^{N} \alpha(\phi_{L})
\label{eq4}
\end{equation}
where $\theta_{L}^{N}$ is the combined feature map, $N$ is number of encoders, $L$ is output block level. $\alpha$ is the angle needed to obtain a $0^{\circ}$ orientation for the feature map $\phi_{L}$ at level L, $\alpha = 360^{\circ} - \beta$ where $\beta$ is the angle of initial rotation.

\section{Experiments}
\subsection{Training Procedure} 
The network follows a modified form of training compared to the base model. The network is trained for 100 epochs with the initial learning rate set to 0.001. This learning rate halves every 20 epochs. The AdaMAX optimiser \cite{kingma2014adam} is used with the $\beta_{1}$ and $\beta_{2}$ hyper-parameters set to 0.9 and 0.999 respectively. The $L_1$ loss function is used for training.

The network is trained on the Vimeo90K dataset which consists of $73,171$ triplets \cite{xue2019video}. $256 \times 256$ crops are taken from the input images of the size $448 \times 256$. Augmentation is applied by using horizontal and vertical flipping as well as switching the temporal order of the input frames for a probability of 0.5.
\subsection {Experimental Setup}
PyTorch \cite{paszke2017automatic} is used to implement this network. The AdaCoF layer is implemented using CUDA and CuDNN \cite{chetlur2014cudnn}. The proposed network is trained on the NVIDIA 2080Ti GPU.
\subsection {Evaluation procedure}
Peak Signal to Noise Ratio (PSNR) and Structural Similarity (SSIM) \cite{wang2004image} are used to quantify performance on the evaluation sets. To provide a fair comparison to the base approach, the Middlebury other dataset \cite{baker2011database} with publicly available ground truth, selected sequences of UCF-101 \cite{soomro2012ucf101} and the DAVIS \cite{perazzi2016benchmark} set are used. Evaluations on the BVI-HFR set \cite{bvihfr} are performed for the proposed method as well as another kernel-based method namely $SepConv++$ \cite{niklaus2021revisiting}. The first 6 videos are taken for evaluation and the frames are arranged in triplets of 3. Each 120fps video sequence consists of 1200 frames. Full-length videos are used for evaluation as this is better linked to real-world uses of interpolation.
\subsection{Ablation study}
In order to evaluate the effectiveness of the proposed model and the impact multiple encoders and rotation have on performance, ablation studies on the Middlebury, UCF-101 and DAVIS sets are conducted. All these networks are trained using the same procedure noted in Subsection 3.1 with the dataset being randomly augmented for each model on the fly. All results presented in this paper are for the kernel size $F=5$ and the kernel dilation is set to 1.

The ablation studies are conducted on the following configurations:

\textbf{\textit{- AdaCoFNet }}: the pre-trained checkpoint released by the authors \cite{lee2020adacof}

\textbf{\textit{- AdaCoFNet Retrained (AdaCoFNet-R)}}: the original AdaCoF model is retrained to allow for a more balanced comparison as all networks would be trained using the same procedure.

\textbf{\textit{- N encoders}}:  $N$ encoders are used as the backbone of the network, the extracted features from the $N$ encoders are combined at each feature level. A skip connection of the combined features is passed to the decoder. This is referred to as $PRNet_{N}$ where $N$ is the number of encoders

\textbf{\textit{- Four encoders + Rotation}}: Same as the four encoder configuration but the input to the first encoder, second, third and fourth encoders is rotated by $0^{\circ}$, $90^{\circ}$, $180^{\circ}$, $270^{\circ}$. Before the features can be combined, the feature maps from each encoder are rotated so that they are the same orientation ($0^{\circ}$). The combined features are then passed to the decoder in the form of a skip connection. This is referred to as $PRNet_{4}^{*}$

\begin{table}[t]
\centering
\arrayrulecolor{black}
\resizebox{\columnwidth}{!}{%
\begin{tabular}{cccccccc} 
\hline
\multirow{2}{*}{\begin{tabular}[c]{@{}c@{}}Number of \\~Parameters~\end{tabular}} & \multirow{2}{*}{Model~}  & \multicolumn{2}{c}{\begin{tabular}[c]{@{}c@{}}\textit{Middlebury}\\\textit{Other}\end{tabular}} & \multicolumn{2}{c}{\textit{UCF-101~}}                                & \multicolumn{2}{c}{\textit{DAVIS~}}                                   \\
                                                                                  &                          & PSNR                             & SSIM                                                         & PSNR                             & SSIM                              & PSNR                             & SSIM                               \\ 
\midrule
1,931,491                                                                         & \textbf{$PRNet_{1}$}     & 35.848                           & 0.9594                                                       & 34.876                           & 0.9646                            & 26.728                           & 0.8084                             \\
2,413,123                                                                         & \textbf{$PRNet_{2}$}     & 35.832                           & 0.9588                                                       & 34.869                           & 0.9643                            & 26.837                           & 0.8115                             \\
2,894,755                                                                         & \textbf{$PRNet_{3}$}     & 36.094                           & 0.9620                                                       & 34.778                           & 0.9645                            & 27.007                           & 0.8112                             \\
3,376,387                                                                         & \textbf{$PRNet_{4}$}     & 36.128                           & 0.9612                                                       & 34.886                           & 0.9645                            & 27.004                           & 0.8151                             \\
3,376,387                                                                         & \textbf{$PRNet_{4}^{*}$} & \textcolor{red}{\textbf{36.258}} & \textbf{\textcolor{blue}{0.9632}}                            & \textcolor{red}{\textbf{34.891}} & \textbf{\textcolor{blue}{0.9649}} & 26.995                           & \textcolor{blue}{\textbf{0.8153}}  \\
21,843,427                                                                        & $AdaCoFNet$ \cite{lee2020adacof}             & 35.692                           & 0.9586                                                       & 34.851                           & 0.9645                           & 26.603                            & 0.8062                             \\
21,843,427                                                                        & \textbf{$AdaCoFNet-R$}   & 35.799                           & 0.9601                                                       & 34.875                           & 0.9641                            & \textcolor{red}{\textbf{27.051}} & 0.8117                             \\
\hline
\end{tabular}
}
\caption{The performance of different network architectures on the Middlebury Other, UCF-101 and DAVIS sets. The results in red are the best PSNR results and those in blue the best SSIM results.}
\end{table}
\begin{table}[t]
\centering
\arrayrulecolor{black}
\resizebox{\columnwidth}{!}{%
\centering
\begin{tabular}{ccccccc} 
\hline
\multicolumn{1}{l}{~}                                                 & $PRNet_{1}$ & $PRNet_{2}$ & $PRNet_{3}$ & $PRNet_{4}$ & $PRNET_{4}*$ & $AdaCoFNet$  \\ 
\midrule
\begin{tabular}[c]{@{}c@{}}Reduction in\\Parameters (\%)\end{tabular} & 91.2        & 89.0        & 86.7        & 84.5         & 84.5         & 0            \\
Runtime (s)                                                           & 0.0302      & 0.0367      & 0.0433      & 0.0498       & 0.0520       & 0.0358       \\
\hline
\end{tabular}}
\caption{The runtime of each model on when interpolating a 480x640 image, using Urban from the Middlebury Evaluation as a test sequence to measure runtime}
\end{table}
\begin{table*}
\centering
\arrayrulecolor{black}

\centering
\resizebox{15cm}{!}{
\begin{tabular}{ccccccccccccc} 
\hline
~           & \multicolumn{2}{c}{\textit{Bobblehead ~}} & \multicolumn{2}{c}{\textit{Books ~}} & \multicolumn{2}{c}{\textit{Bouncyball ~}} & \multicolumn{2}{c}{\textit{Catch ~}} & \multicolumn{2}{c}{\textit{Catch-track ~}} & \multicolumn{2}{c}{\textit{Average ~}}  \\
            & PSNR   & SSIM                             & PSNR   & SSIM                        & PSNR   & SSIM                             & PSNR   & SSIM                        & PSNR   & SSIM                              & PSNR   & SSIM                           \\ 
\midrule
PRNet4*     & 21.536 & 0.5666                           & 39.288 & 0.9649                      & 34.956 & 0.9216                           & 38.345 & 0.9826                      & 31.426 & 0.8323                            & 33.110 & 0.8536                         \\
AdaCoFNet-R & 21.519 & 0.5635                           & 38.947 & 0.9624                      & 34.894 & 0.9208                           & 38.290 & 0.9825                      & 31.346 & 0.8317                            & 32.999 & 0.8522                         \\
SepConv++ \cite{niklaus2021revisiting}  & 21.565 & 0.5686                           & 40.601 & 0.97151                     & 35.429 & 0.9262                           & 38.394 & 0.9827                      & 30.872 & 0.8241                            & 33.372 & 0.8546                         \\
\hline
\end{tabular}}
\caption{The PSNR and SSIM results on an evaluation of the first 5 sequences of the BVI-HFR set}
\end{table*}
\begin{figure*}
\centering
\begin{center}
   \includegraphics[width=0.9\linewidth]{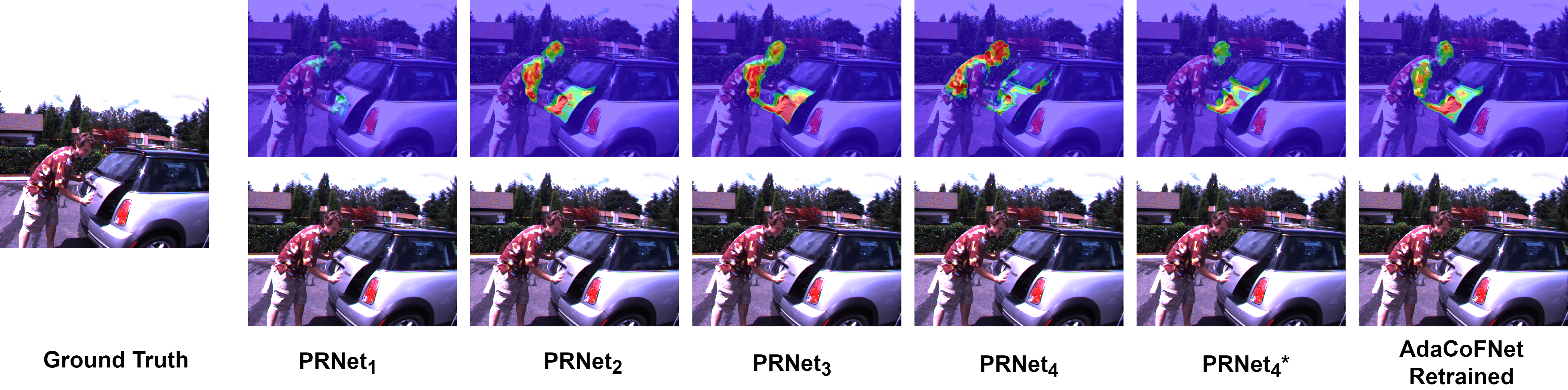}
\end{center}
   \caption{A visualisation of the highest-scoring attention map for the models used in the ablation. Red indicates areas where there are high attention, blue indicates areas of low attention }
\label{fig3:long}
\label{fig3:legacy}
\end{figure*}

In our ablation, we first start by investigating the impact the training setup has on model performance. The results from the publicly available checkpoint of $AdaCoFNet$ in all three datasets are lower than that of the retrained model using our setup. Our retrained model performs 0.107dB,0.447dB and 0.0240dB more on the Middlebury Other, UCF-101 and DAVIS sets.

The impact of increasing the number of encoders and applying rotation can be observed in Table 1. The impact is most notable on the Middlebury Other set where the PSNR and SSIM increased by 0.41dB and 0.0038 when comparing the performance of $PRNet_{1}$ to $PRNet_{4}^{*}$. The impact of rotations on performance can be observed by observing the difference in performance between $PRNet_{4}$ and $PRNet_{4}^{*}$ where there is a 0.13dB increase in PSNR on the Middlebury Other set. The performance on the UCF-101 and DAVIS sets remains close to the $PRNet_{4}$ indicating that rotation might be more beneficial in cases where overall PSNR performance is higher (Middlebury) than those with lower PSNR (UCF-101, DAVIS). 
The results of this ablation indicate that the 4-encoder model with rotation $PRNet_{4}^{*}$ obtains the best performance on the Middlebury Other (0.459dB more, 0.0031 more for SSIM) and similar performance levels on the UCF-101 (0.016dB more, 0.0008 more for SSIM) and DAVIS (0.056dB less, 0.0036 more for SSIM). 

This performance, however, comes at a cost. When interpolating a 640x480 frame, using the Urban sequence from the Middlebury evaluation set, $PRNet_{4}$ and $PRNet_{4}^{*}$ take 0.0272s and 0.0325s more to interpolate the frame than $AdaCoFNet$ as shown in Table 2. $PRNet_{4}^{*}$ takes slightly longer than $PRNet_{4}$ in interpolating the Urban sequence due to the presence of the rotation operation at different parts of the network. Despite the reduced parameters, runtime increases by approximately 0.005s for each encoder added and this can be explained by the fact that the input images have to be passed to all 4 encoders. Also, convolution operations for the 4 encoders have to be computed as opposed to one set of convolutions for the standard architecture. This is also a reason for the slower runtime of the PRNet models compared to the original. However, from a memory perspective, the PRNet models use less VRAM.

This variation in performance between the various models in the ablation can be explained using the occlusion maps presented in Figure 2. The blue colour indicates occluded pixels in frame 1 and the red occluded pixels in frame 2, green indicates no occlusion. The occlusions are visualised using the same procedure as \cite{lee2020adacof}. When observing the first row, it can be seen that all the models manage to detect occlusions in the areas around the hands and the juggling balls. What differs is the extent to which occluded pixels are detected and this can be observed through the difference in how much blue and red there are around the main areas of occlusion. 
For the second row featuring moving vehicles, the difference in occlusion detection can be observed especially in the top left corner. $PRNet_{1}$ does not detect any occlusion in the top left and top right of the image, whereas the rest of the models detect this to varying degrees. AdaCoFNet retrained does not detect much occlusion in these areas compared to $PRNet_{3}$, $PRNet_{4}$ and $PRNet_{4}^{*}$. 

A similar case can be observed in the third row of images which shows a forested area. $PRNet_{1}$ and $AdaCoFNet-R$ barely detect much occlusion in the top left of the image. In comparison, the occluded areas detected in the top left can be seen in $PRNet_{3}$, $PRNet_{4}$ and $PRNet_{4}^{*}$. 

This indicates that occlusion has a role to play in image synthesis and the performance of the models.
Another aspect to consider is the visualisation of the attention maps of the networks. Figure 3 shows the attention maps obtained from the last upsampling layer before the kernel generation and AdaCoF operation. The attention map is superimposed on the output interpolated image of the models. Similar to occlusion maps, each models focuses on the areas of high motion but what differs is the extent of the attention on these areas. $PRNet_{4}$, $PRNet_{4}^{*}$ and $AdaCoFNet-R$ have a high attention on these areas of high motion and cover a larger area compared to $PRNet_{1}$, $PRNet_{2}$ and $PRNet_{3}$. Another aspect to note is how much attention is given to the shutter, with $PRNet_{4}$ being the only network that focused on a significant portion of the shutter. The difference in attention could also play a role in the synthesis of the output frame. It is recommended to view the supplementary material for more occlusion and attention map visualisations as well as comparisons of interpolated image output.

After selecting $PRNet_{4}^{*}$ as our preferred model, evaluations are conducted on the BVI-HFR videos \cite{bvihfr}, shown in Table 3. A comparison with $SepConv++$  \cite{niklaus2021revisiting} is provided to show possible benefits of applying the paper's proposed modifications to our method. Though a fairer comparison would be to retrain $SepConv++$ on the entire Vimeo-90K dataset as $SepConv++$ is only trained on the train set. The reason $SepConv++$ is selected for comparison is to show the potential of applying these improvements to our network. The $SepConv++$ paper applies modifications to an earlier kernel-based network $SepConv$ \cite{niklausiccv2017} and shows that with these modifications, performance could be comparable to state-of-the-art. 

As seen in Table 3, $PRNet_{4}^{*}$ performs better on all sequences compared to $AdaCoFNet-R$. $SepConv++$ performs better compared to $PRNet_{4}^{*}$. As $AdaCoFNet$ and $SepConv++$ are both kernel-based networks that are based on the principle of using a UNet to extract weights for a special convolution operation, applying some of the modifications introduced in \cite{niklaus2021revisiting} might result in better performance. There is definitely scope for improving $SepConv++$ uses a memory inefficient kernel size of 51 but $AdaCoFNet$ remedies these issues. Also, $PRNet_{4}^{*}$ uses less parameters than $SepConv++$ (3.7m vs 14m). Thus, applying these improvements to $PRNet_{4}^{*}$ might be a possible route.

Experiments are also conducted to compare the memory usage and runtime of the best performing PRNet models ($PRNet_{4}$ and $PRNet_{4}^{*}$) to $AdaCoFNet$ and $SepConv++$ on full-length videos of different resolutions. This would help with investigating the practicality of our method compared to others in real-time. The publicly implementation for $SepConv++$ is used for this evaluation. The evaluation is conducted on the NVIDIA Quadro RTX 5000. This is done to allow for experiments on the 4096x2048 resolution which is becoming more popular with users. The CPU used on the machine is the Intel Xeon 5218R.

\begin{table*}
\centering
\resizebox{14cm}{!}{
\begin{tabular}{cccccc} 
\toprule
Video Resolution           & Runtime Information                            & $PRNet_{4}^{*}$ & $PRNet_{4}$  & $AdaCoFNet$ & $SepConv++$  \\ 
\midrule
\multirow{2}{*}{4096x2160} & Memory Usage (MB)                 & ~13953  & ~13715  & ~14421    & 15633      \\
                           & Average Interpolation Speed (s/f) & 1.45    & 1.39    & ~0.965    & ~2.81      \\ 
\midrule
\multirow{2}{*}{2048x1080} & Memory Usage (MB)                 & 4197    & 4139    & ~4389     & ~5081      \\
                           & Average Interpolation Speed (s/f) & ~0.357  & ~0.344  & ~0.293    & ~0.694     \\ 
\midrule
\multirow{2}{*}{1280x720}  & Memory Usage (MB)                 & 2297    & ~2297   & ~2443     & ~2697          \\
                           & Average Interpolation Speed (s/f) & ~0.157  & ~0.152  & ~0.105    & ~0.290          \\ 
\midrule
\multirow{2}{*}{640x360}   & Memory Usage (MB)                 & ~1295   & 1295~   & ~1413     & ~1413      \\
                           & Average Interpolation Speed (s/f) & ~0.0414 & ~0.0398 & ~0.0286   & ~0.0735    \\ 
\midrule
\multirow{2}{*}{320x180}   & Memory Usage (MB)                 & ~1035   & 1035~   & ~1097     & ~1065      \\
                           & Average Interpolation Speed (s/f) & ~0.0113 & ~0.0104 & ~0.00891  & ~0.0201    \\
\toprule
\end{tabular}}
\caption{The average interpolation speed (seconds/frame) and GPU memory usage (MB) of $PRNet_{4}$, $PRNet_{4}^{*}$, $AdaCoFNet$ and $SepConv++$. }
\end{table*}

For the 4096x2160 resolution, $PRNet_{4}$ uses the least amount of GPU memory (13715MB) followed by $PRNet_{4}^{*}$ (238MB more), $AdaCoFNet$ (706MB more) and $SepConv++$ (1918MB more). The trends relating to the average interpolation speed  mirror those in Table 2 with $PRNet_{4}^{*}$ taking longer than $PRNet_{4}$ and both these methods taking longer than $AdaCoFNet$. $SepConv++$ takes the longest amount of time (2.81s), approximately 2.9 times how long it takes $AdaCoFNet$ to synthesise a frame. $PRNet_{4}^{*}$ takes approximately 1.5 times longer than $AdaCoFNet$ for the same operation.

For the 2040x1080 resolution, the difference in GPU memory usage becomes less stark with $PRNet_{4}$ still using the least amount of GPU memory (4139) followed by $PRNet_{4}^{*}$ (58MB more), $AdaCoFNet$ (250MB more) and $SepConv++$ (692MB). The difference in runtime has also decreased with $PRNet_{4}$, $PRNet_{4}^{*}$ and $SepConv++$ taking 1.22, 1.17 and 2.37 times longer than AdaCoFNet.

For the resolutions of 1280x720, 640x360 and 320x180, the memory usage of $PRNet_{4}$ and $PRNet_{4}^{*}$ becomes the same whilst still being lower than that of $AdaCoFNet$. This indicates that at lower resolutions, rotations are less demanding from a memory usage perspective. At the resolutions of 640x360 and 320x180, $SepConv++$ uses a similar amount of memory to other methods.

What can be concluded from Table 4 is that although $PRNet_{4}^{*}$ takes between  1.17-1.5 times longer to synthesise a frame, the model uses less GPU memory and this would be of most benefit at higher resolutions where the gulf in memory usage between $AdaCoFNet$ and $PRNet_{4}^{*}$ widens. At all the different resolutions, the trends in runtime were similar with $SepConv++$ taking the longest, followed by $PRNet_{4}^{*}$, then $PRNet_{4}$ and then $AdaCoFNet$. 

Another aspect highlighted by this evaluation is the possible impact on runtime applying some of the modifications of $SepConv++$ to $AdaCoFNet$ discussed before. At all resolutions, $SepConv++$ took the longest. However, as observed in Table 3, it also performs better than $PRNet_{4}^{*}$ and $AdaCoFNet$ on the BVI-HFR set so the issue whether performance is more important than runtime needs to be weighed. As memory usage and runtime are both important considerations, the benefits of the methods listed need to be weighed up. For those with limited memory, $PRNet_{4}$ and $PRNet_{4}^{*}$ are good candidates. However, for those who would like faster frame synthesis, using a smaller PRNet such as $PRNet_{2}$ or $PRNet_{3}$ with comparable performance to $AdaCoFNet$ might be a good option. 

\section{Conclusion}
This paper has presented a new approach to reduce network parameters and memory usage by utilising multiple encoders to compensate for the removal of the deeper, more intrinsic features that tend to be present with deeper networks. Through conducting ablations to determine which model gives the best performance relative to parameter count, we have demonstrated that in fact this multi-encoder approach is just as effective, if not better than $AdaCoFNet$ in some cases whilst using 84.5$\%$ less parameters. This was further verified through testing on the BVI-HFR dataset \cite{bvihfr} whereby both the proposed model and $AdaCoFNet$ obtained similar results, in some cases, our proposed model even obtained a higher PSNR.
For future work, it would be interesting to investigate the impact of kernel size and dilations on model performance. Another aspect to consider would be applying our method to other kernel-based networks and UNet architectures in other fields to see whether similar benefits could be observed. As our approach is applied to the UNet part of the architecture, is easy to implement and follows a similar training procedure to the baseline, this is a potential avenue to consider.

\textbf{Acknowledgements:} This work was supported under the iCase grant (Ref: 2246814) from the Engineering and Physical Sciences Resarch Council (EPSRC) in collaboration with the British Broadcasting Cooperation (BBC)

{\small
\bibliographystyle{ieee_fullname}
\bibliography{egbib}
}

\end{document}